\documentclass{article}
\usepackage{spconf,amsmath,graphicx,hyperref}
\usepackage{cite}
\usepackage{amsmath,amssymb,amsfonts}
\usepackage{algorithmic}
\usepackage{graphicx}
\usepackage{textcomp}
\usepackage{xcolor}
\def\BibTeX{{\rm B\kern-.05em{\sc i\kern-.025em b}\kern-.08em
    T\kern-.1667em\lower.7ex\hbox{E}\kern-.125emX}}

\title{DynaPose4D: High-Quality 4D Dynamic Content Generation via Pose Alignment Loss}
\name{Jing Yang  \qquad Yufeng Yang}
\address{Sun Yat-sen University}

\begin{document}
\maketitle
\begin{abstract}

\end{abstract}
Recent advancements in 2D and 3D generative models have expanded the capabilities of computer vision. However, generating high-quality 4D dynamic content from a single static image remains a significant challenge. Traditional methods have limitations in modeling temporal dependencies and accurately capturing dynamic geometry changes, especially when considering variations in camera perspective. To address this issue, we propose DynaPose4D, an innovative solution that integrates 4D Gaussian Splatting (4DGS) techniques with Category-Agnostic Pose Estimation (CAPE) technology. This framework uses 3D Gaussian Splatting to construct a 3D model from single images, then predicts multi-view pose keypoints based on one-shot support from a chosen view, leveraging supervisory signals to enhance motion consistency. Experimental results show that DynaPose4D achieves excellent coherence, consistency, and fluidity in dynamic motion generation. These findings not only validate the efficacy of the DynaPose4D framework but also indicate its potential applications in the domains of computer vision and animation production.

\section{Introduction}
\label{sec:intro}
Implicit neural rendering techniques have been widely adopted for various tasks, such as pose and shape estimation, novel view synthesis (NVS), and static 3D or dynamic 4D generation . Among these, 4D Gaussian Splatting (4DGS) has pioneered the generation of dynamic scenes from single images or video sequences by enforcing strict temporal consistency across frames\cite{li2021neuralsceneflowfields,Z2}\cite{li2023dynibarneuraldynamicimagebased,Z1}. Despite its progress, 4D Gaussian Splatting still faces limitations, particularly in handling temporal sequence dependencies and dynamic changes. Issues arise when dealing with changes in camera perspectives, which lead to insufficient visual consistency and hinder the generation of long-duration, complex 3D motions. Consequently, dynamic scenes generated using these methods may lack coherence, appearing unnatural or inconsistent, which impairs the transition from static to dynamic content.
\begin{figure}[t]
    \centering
    \includegraphics[width=1.0\linewidth]{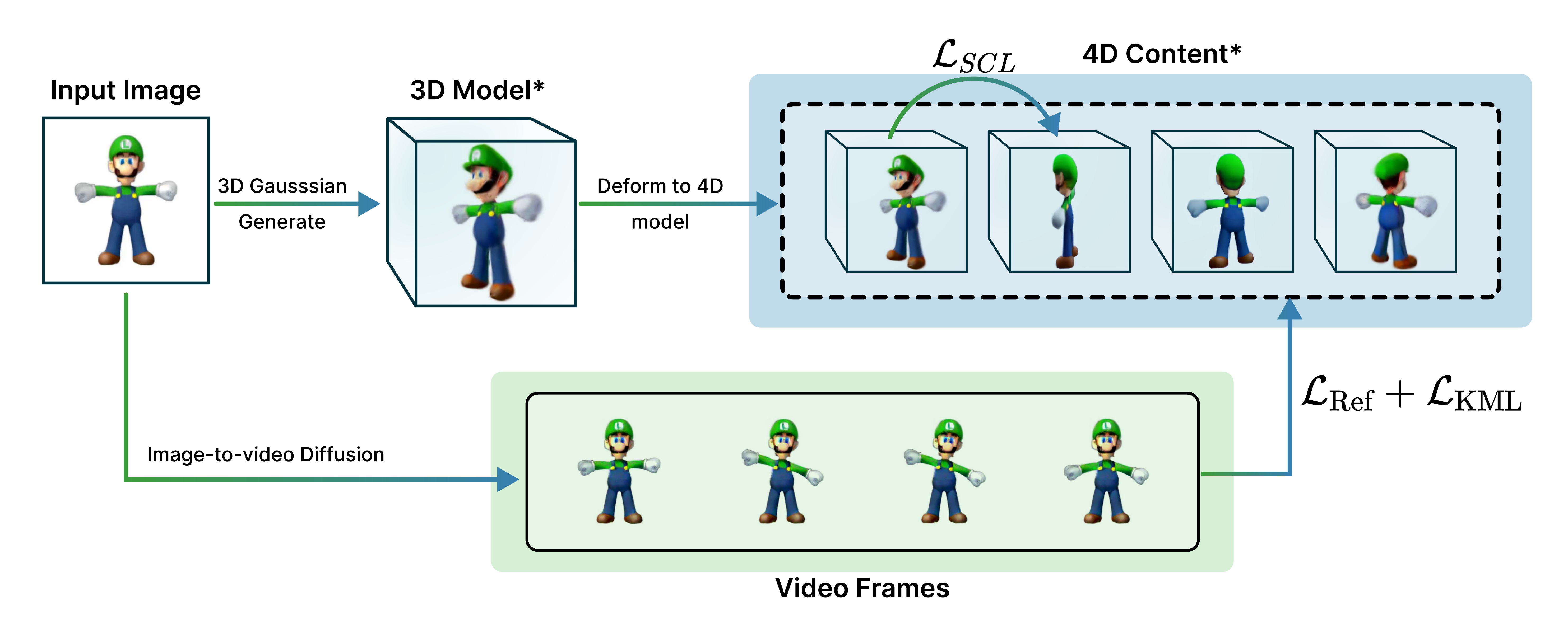}
\caption{\textbf{Overview of DynaPose4D.} 
From a single image, 3D Gaussians are deformed into 4D content. Temporal consistency is enforced by $\mathcal{L}_{\text{SCL}}$, while $\mathcal{L}_{\text{Ref}}+\mathcal{L}_{\text{KML}}$ supervise visual consistency and refine pose transitions.}
    \label{fig:framework}
\end{figure}

To address these challenges, we propose the DynaPose4D, which integrates 4D Gaussian Splatting (4DGS) techniques with Category-Agnostic Pose Estimation (CAPE) technology\cite{xiong2023capecameraviewposition}. The framework, shown as Fig. \ref{fig:framework}, begins by using 3D Gaussian Splatting to construct a 3D model from a single static view. CAPE is then employed to extract pose keypoints, utilizing CNNs for image feature extraction and GCNs for graph-structured data, thereby improving the accuracy of pose pattern detection in images. This allows precise pose keypoints to be extracted from dynamic objects in video sequences, which serve as supervisory signals to ensure smooth transitions between static and dynamic content.

In this research, we make several key contributions:
\begin{enumerate}
\item We introduce DynaPose4D, a 4D content generation framework that supports multimodal inputs, including single image and video sequences.
\item Involving the use of advanced pose estimation techniques, DynaPose4D can accurately infer the position of keypoints and their temporal aspects for dynamic objects in any given image or video.
\item We propose a novel method that uses pose keypoints as conditions to guide 4D video generation, ensuring a high level of spatio-temporal consistency of the generated 4D content, while accurately preserving the keypoint trajectories.
\end{enumerate}
Experimental results demonstrate that DynaPose4D achieves significant improvements in dynamic objects generation, both for single images and video sequences. We observe that the performance of the model under dynamic objects is significantly improved by the accurate detection of CAPE\cite{b6}\cite{b12}.

\section{Prior work}
Generating high-quality 4D dynamic content from a single static image integrates challenges from multiple fields in computer vision, including 3D reconstruction, dynamic scene modeling, and pose estimation. Traditional 3D reconstruction methods, such as Neural Radiance Fields (NeRF)\cite{mildenhall2020nerfrepresentingscenesneural}, have made significant strides in novel view synthesis by representing scenes as volumetric radiance fields. However, NeRF requires multiple views for training, limiting its effectiveness for single-image reconstruction. Zero-1-to-3 \cite{liu2023zero1to3zeroshotimage3d}addresses this issue by leveraging large-scale diffusion models to predict novel views from a single image but is restricted to static scenes, lacking the ability to model temporal dynamics.

In dynamic scene generation and 4D modeling, methods like Dynamic NeRFs and 4D Gaussian Splatting (4DGS) extend NeRF to incorporate time as an additional input, allowing for the modeling of temporal changes. While these methods enable dynamic scene generation, they still rely on multi-view and multi-frame data for training. Even with advancements like DreamGaussian4D, which enforces temporal coherence, accurately capturing dynamic geometry and temporal dependencies remains challenging, particularly when dealing with varying camera perspectives, resulting in artifacts and inconsistencies in generated content.

Pose estimation plays a critical role in understanding and generating motion within dynamic scenes. Techniques like OpenPose\cite{cao2019openposerealtimemultiperson2d} and SMPL\cite{SMPL} have laid the groundwork for 2D and 3D pose estimation, while category-agnostic frameworks like PoseAnything have improved keypoint localization across diverse object categories. Generative models, such as Text2Video-Zero\cite{khachatryan2023text2videozerotexttoimagediffusionmodels}, utilize pose information to guide video generation, yet they often depend on multi-view or pose data inputs during inference, leaving gaps in their ability to generate dynamic 4D content directly from single images.

\section{Method}

Here, we introduce our proposed dynamic 4D Gaussian Splatting generation process and a pose-supervision loss designed to enforce visual consistency during training. First, we review the process of generating 3D models from a single image, followed by an analysis of the transition from static 3D models to dynamic 4D content. Finally, we present a pose-supervision loss to ensure high-quality and coherent motion generation.

\subsection{Neural Rendering for 4D Content}

\textbf{Static 3D generation from single images. }For the image-to-3D task, we employ the Zero-1-to-3 model, which uses a single RGB image to generate novel 3D viewpoints by leveraging geometric priors learned from large-scale diffusion models. The model predicts different perspectives of the object by conditioning on camera transformations, enabling high-quality 3D reconstructions from a single view.

\textbf{Image-to-video diffusion for dynamic prior. }We employed the Stable Video Diffusion (SVD) model\cite{blattmann2023stablevideodiffusionscaling} to generate the transformation process from a single input image to a driving video. SVD is a diffusion-based generative model initially used to create high-quality static images. To extend this model for video generation tasks, we leveraged its temporal dependency by inputting a single image and introducing random noise \(\epsilon\), generating a time-sequence video. This process can be described by the following equation:

\begin{equation}
\left\{I^{\text{Ref}}\right\}_{\tau=1}^T = f_\psi\left(\epsilon ; I^{\text{Sup}}\right),
\end{equation}
where \(I^{\text{Sup}}\) represents the input image, \(\left\{I^{\text{Ref}}\right\}_{\tau=1}^T\) is the generated video sequence, \(\epsilon\) is the random noise, \(f_\psi\) is the image-to-video diffusion model, and \(T\) is the number of time steps in the video. Through this model, we can generate a driving video containing dynamic motion information from a single input image.

\begin{figure}
    \centering
    \includegraphics[width=\linewidth]{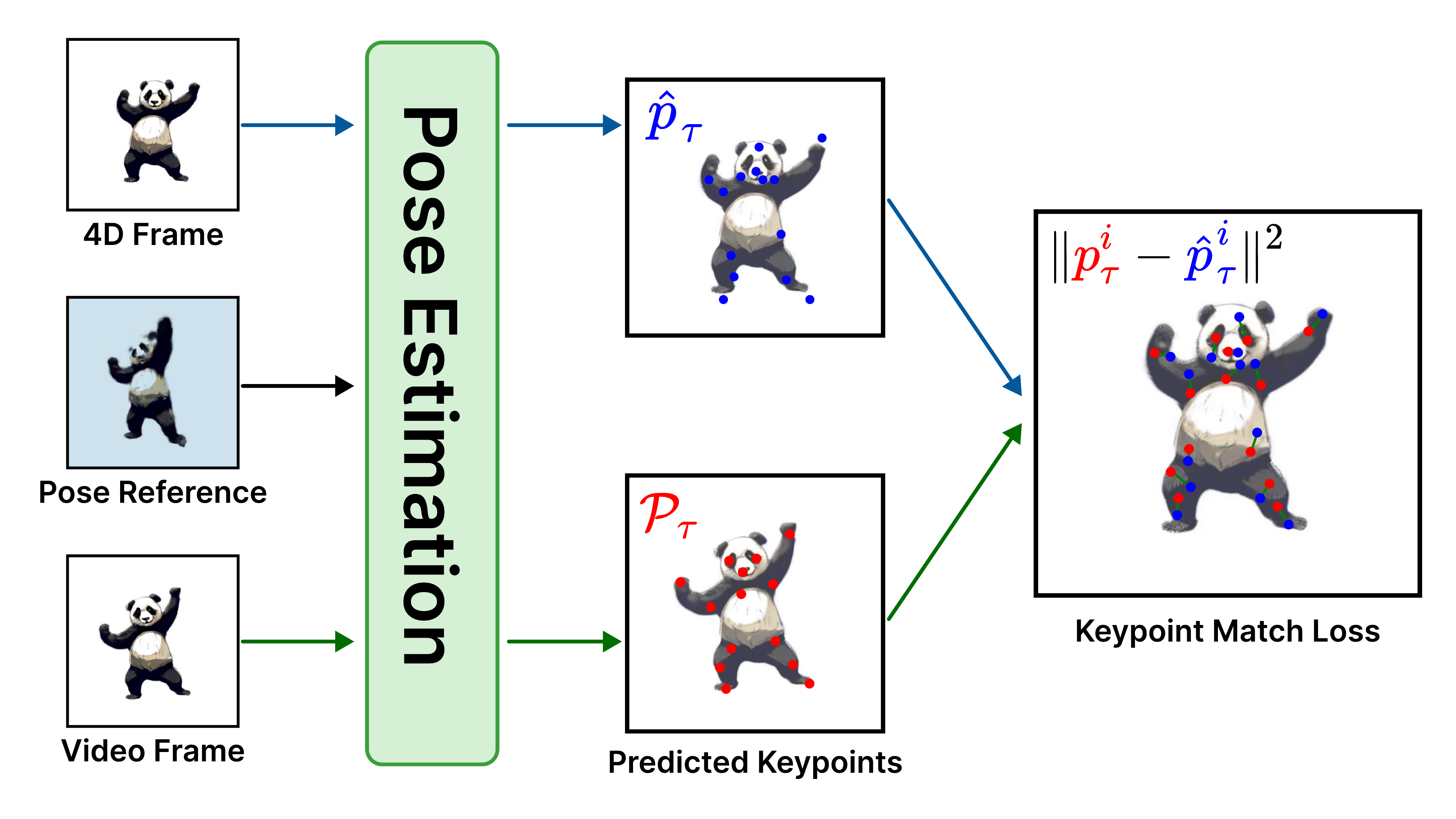}
    \caption{KML minimizes MSE between poses extracted from rendered and predicted frames under one-shot support.
}
    \label{fig:loss}
\end{figure}
\textbf{Dynamic 4DGS generation. }Next, we use 4D Gaussian Splatting (4DGS) to extend the static 3D model into dynamic 4D content\cite{pumarola2020dnerfneuralradiancefields}\cite{park2021hypernerfhigherdimensionalrepresentationtopologically}. 4DGS explicitly models dynamic changes in both spatial and temporal dimension, predicting spatial position, rotation, and scaling variations while ensuring temporal consistency of the generated content. The deformation process can be formulated as:
\begin{equation}
S_\tau^{\prime} = \phi(S, \tau),
\end{equation}
where $\phi$ is the deformation network, $S$ represents the static 3D Gaussians, $\tau$ denotes the time step, and $S^{\prime}_\tau$ is the deformed 3D Gaussians at time $\tau$. To further enhance the quality of the generated dynamic content, we take Score Distillation Sampling (SDS) method for efficient initialization, while optimize the deformation process by minimizing the mean squared error (MSE) between the rendered result of 4DGS and the reference frames from SVD:
\begin{equation}
\mathcal{L}_{\text{Ref}} = \frac{1}{T} \sum_{\tau=1}^T \left\| f\left(S_\tau^{\prime}, o^{\text{Ref}}\right) - I_\tau^{\text{Ref}} \right\|_2^2,
\end{equation}
where $I_\tau^{\mathrm{Ref}}\in\left\{I^{\text{Ref}}\right\}_{\tau=1}^T$ is the reference image at time $\tau$, $\mathcal{L}_{\mathrm{Ref}}$ is the MSE loss for the reference viewpoint $o_\text{Ref}$ which aligns to $I_\tau^{\text{Ref}}$, $T$ is the number of time steps, and $f\left(S^{\prime}_\tau, o^{\text{Ref}}\right)$ is the rendering function of the deformed Gaussians.

\subsection{Pose Alignment Loss}

To further enhance the quality and coherence of the generated motion while ensuring alignment with the input keypoint poses, we introduce a pose supervision losscite\cite{cao2017realtimemultiperson2dpose,Z4}\cite{fang2018rmperegionalmultipersonpose,Z3}, illustrated as Fig. \ref{fig:loss}. This loss guides the dynamic transformation of the 3D Gaussians, ensuring the generated motion achieves high spatio-temporal consistency and matches the pose keypoints in the driving video frames. Specifically, given the input static image and it's pose keypoints, we use PoseAnything\cite{Hirschorn_2024,Z8} to predict pose keypoints from each $I_{\tau}^{\text{Ref}}$ and $f(S_\tau^{\prime}, o^{\text{Ref}})$ with support of one-shot pose keypoints in $I^{Sup}$, denote as $p_\tau\in\mathbb{R}^{N,2}$ and $\hat p_\tau\in\mathbb{R}^{N,2}$, respectively. The Keypoint Match Loss (KML) can be formulated as:
\begin{equation}
\mathcal{L}_{\text{KML}} = \frac{1}{N} \sum_{i=1}^{N} \sum_{\tau=1}^{T-1} \| p_\tau^i - \hat{p}_\tau^i \|^2
\end{equation}
Furthermore, to ensure the origins of the generated 3D Gaussians $\mathcal{P}_\tau\in \mathbb{R}^{M, 3}$ maintains smoothness and temporal consistency, we define a Spatio-temporal Consistency Loss (SCL)\cite{tulyakov2017mocogandecomposingmotioncontent}\cite{Z7}. This loss prevents abrupt changes in the origins of the Gaussians between consecutive time steps.
\begin{equation}
\mathcal{L}_{\text{SCL}} = \frac{1}{M} \sum_{i=1}^{M} \sum_{\tau=1}^{T-1} \| \mathcal{P}^i_{\tau+1} - \mathcal{P}^i_\tau \|^2
\end{equation}
Here, $N$ indicate the total number of keypoints, while $M$ represents the total number of 3D Gaussians. In summary, we define the Pose Alignment Loss (PAL) as the weighted sum of KML and SCL.

\section{Experiments}
We conducted extensive experiments to evaluate the effectiveness of the proposed DynaPose4D framework in generating high-quality 4D dynamic content from single images. All experiments were performed on a single NVIDIA RTX 3090 GPU with 24 GB of memory.
\textbf{Implementation Details}
We utilized the open-source repository DreamGaussian4D~\cite{b9} as the base framework for 4D Gaussian Splatting to generate dynamic 3D models. For pose supervision, we employed PoseAnything, where $N=14$, to infer subsequent dynamic pose keypoints from a single static image. This provided supervisory signals to enhance motion consistency and temporal coherence in the generated 4D content.
To optimize the deformation process and further enhance the dynamic content of the motion, we used the Mean Squared Error (MSE) loss. We ran 500 iterations with a batch size of 16 to ensure model stability during the later stages of training. We initialized $M = 512$ control Gaussians uniformly within a sphere at a fixed radius of 2. The azimuth angles were sampled uniformly in the range $[-180^\circ, 180^\circ]$. We used consistent scaling parameters throughout the training process. The parameter $T_{\text{max}}$ was linearly decayed from 0.98 to 0.02 over the iterations to facilitate smooth deformation.
\begin{figure*}
    \centering
    \includegraphics[width=\linewidth]{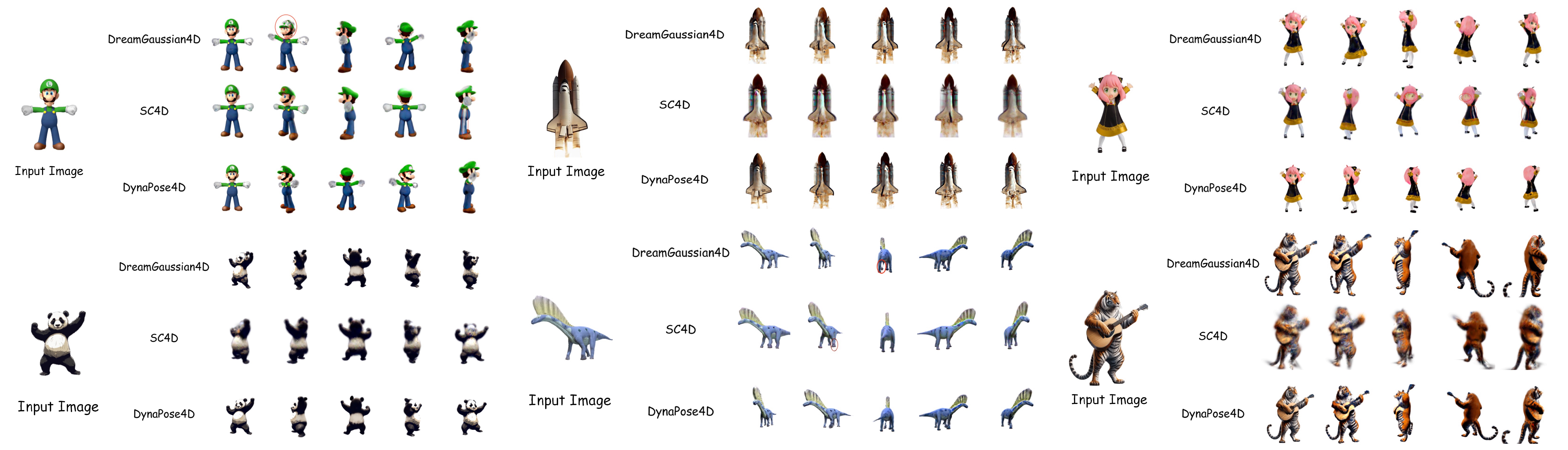}
    \caption{\textbf{visual comparisons among different methods.} DynaPose4D produces more coherent and realistic dynamic content, with better temporal consistency and motion smoothness.}
    \label{fig:compare}
\end{figure*}
\subsection{Evaluation Metrics}
To assess the quality of the generated 4D dynamic content, we conducted evaluations on two publicly available datasets: Consistent4D~\cite{jiang2024consistentd} and Animate124~\cite{zhao2024animate124animatingimage4d}. We employed three widely used metrics to quantitatively evaluate the results:
\textbf{Peak Signal-to-Noise Ratio (PSNR):}\cite{PSNR} Measures the pixel-wise differences between the generated images and ground truth, with higher values indicating better quality.
\textbf{Structural Similarity Index Measure (SSIM):}\cite{SSIM} Assesses the structural similarity and perceptual quality between images, focusing on luminance, contrast, and structure. \textbf{Learned Perceptual Image Patch Similarity (LPIPS):}\cite{zhang2018unreasonableeffectivenessdeepfeatures} Utilizes deep neural network features to evaluate perceptual differences, particularly effective for assessing the coherence and visual quality of dynamic content.
\subsection{Comparative Analysis}
To validate the effectiveness of the proposed DynaPose4D framework, we compared our method with two state-of-the-art open-source methods: DreamGaussian4D~\cite{ren2024dreamgaussian4dgenerative4dgaussian} and SC4D~\cite{wu2024sc4dsparsecontrolledvideoto4dgeneration}. All methods were trained on the Consistent4D and Animate124 datasets using their official codes and default settings.
\begin{table}[htbp]
\caption{Quantitative comparison of different methods on Consistent4D and Animate124 datasets.}
\centering
\begin{tabular}{lcccc}
\hline
\textbf{Method} & \textbf{PSNR}$\uparrow$ & \textbf{SSIM}$\uparrow$ & \textbf{LPIPS}$\downarrow$ \\
\hline
DreamGaussian4D & 18.980  & 0.797  & 0.206  \\
SC4D            & 18.164  & 0.805  & 0.209  \\
\textbf{DynaPose4D}  & \textbf{22.761} & \textbf{0.863} & \textbf{0.122} \\
\hline
\end{tabular}
\label{table:comparison}
\end{table}
Table~\ref{table:comparison} summarizes the quantitative results of the comparison. DynaPose4D achieves superior performance across all metrics, demonstrating significant improvements in both fidelity and perceptual quality.
\subsection{Ablation Study}

To assess the contribution of the pose supervision component in DynaPose4D, we conducted an ablation study by removing the pose supervision, denoted as DynaPose4D w/o pose supervision. We evaluated the model's performance under this configuration to quantify the impact of pose supervision on the quality of the generated content.

\begin{table}[htbp]
\caption{Ablation study on the impact of pose supervision.}
\centering
\begin{tabular}{lccc}
\hline
\textbf{Method} & \textbf{PSNR}$\uparrow$ & \textbf{SSIM}$\uparrow$ & \textbf{LPIPS}$\downarrow$ \\
\hline
\textbf{+Support*} & \textbf43.0244 & \textbf0.9981 &\textbf0.0039 \\
-Support* & \textbf{40.5138} & \textbf{0.9956} & \textbf{0.0048} \\
\hline
\end{tabular}
\label{table:ablation}
\end{table}
Table~\ref{table:ablation} reports the quantitative results of the ablation study. The results indicate that removing the pose supervision mechanism leads to a decrease in performance, particularly in terms of temporal coherence and motion smoothness in long-duration dynamic content. This demonstrates that pose supervision provides critical guidance for the model to capture temporal variations and maintain spatial consistency.
\begin{figure}[htbp!]
    \centering
    \includegraphics[width=1.0\linewidth]{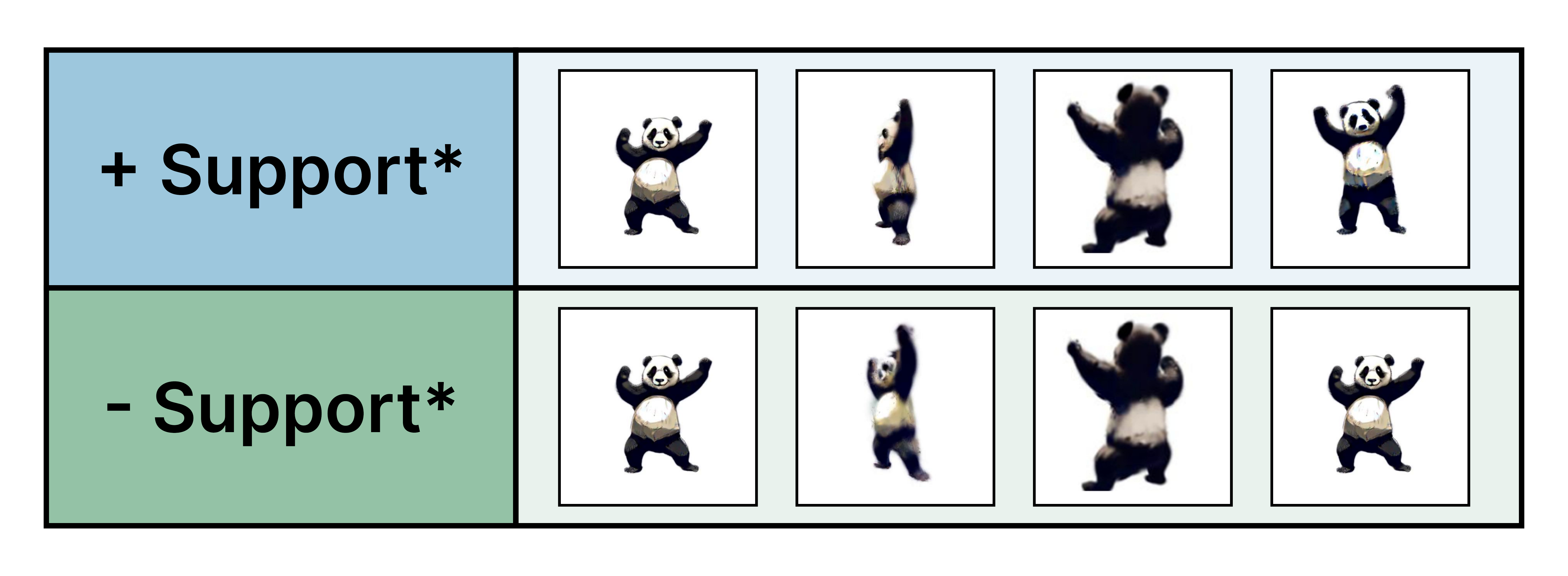}
    \caption{\textbf{Result of ablation studies.} Pose keypoint supervision enhances 4D visual consistency and yields 3D geometry better aligned with the source image.}
    \label{fig:xiaorong}
\end{figure}
\subsection{Discussion}

Fig.~\ref{fig:xiaorong} illustrates visual examples from the ablation study. The images show that without pose supervision, the generated content exhibits artifacts, temporal jitter, and spatial inconsistencies, whereas the proposed DynaPose4D maintains higher-quality and more coherent motion. The experimental results further validate the effectiveness of the framework: by integrating pose estimation with CAPE technology, our method directly tackles the challenge of generating spatio-temporally consistent 4D content from single images. In particular, pose supervision guides the model to infer temporal dynamics more accurately and to preserve fine-grained spatial structure even in challenging scenes with complex motions or self-occlusion. These improvements are not only visible qualitatively but are also supported by quantitative metrics, demonstrating that DynaPose4D achieves more stable and reliable results. Overall, the ablation study highlights that pose supervision is not merely an auxiliary component but a fundamental factor in ensuring robustness and generalization.

\section{Conclusion}
This paper presents \textbf{DynaPose4D}, a framework that generates high-quality 4D dynamic content from a single image by integrating 4D Gaussian Splatting with pose supervision for spatio-temporal consistency. Experiments show clear improvements over state-of-the-art methods in PSNR, SSIM, and LPIPS, confirming both visual fidelity and quantitative gains. Ablation studies highlight pose supervision as crucial, since its removal degrades temporal coherence and motion smoothness. Overall, DynaPose4D effectively captures dynamic changes while preserving spatial consistency, offering a robust solution for challenging scenarios and strong potential for applications such as animation, AR/VR content creation, and motion-driven 3D reconstruction, while also opening opportunities for future research on spatio-temporal generative modeling.

\newpage
\bibliographystyle{IEEEbib}
\bibliography{strings,refs}

\end{document}